\documentclass[USenglish,twocolumn]{article}
\usepackage[utf8]{inputenc}
\usepackage[big,online]{dgruyter}
\usepackage[sort&compress,square,numbers]{natbib}
\usepackage{times}
\usepackage{hyperref}       
\usepackage{siunitx}
\usepackage{color}

\begin{document}

  \journalname{Current~Directions~in~Biomedical~Engineering}
  \journalyear{2018}
  \journalvolume{4}
  \journalissue{1}
  \startpage{1}
  \DOI{10.1515/cdbme-2018-XXXX}
  \openaccess

\title{OmniPD: One-Step Person Detection in Top-View Omnidirectional Indoor Scenes}
\runningtitle{OmniPD}

\author[1]{Jingrui Yu}
\author[2]{Roman Seidel}
\author[3]{Gangolf Hirtz} 
\runningauthor{J.Yu et al.}

\affil[1]{\protect\raggedright 
  Faculty of Electrical Engineering and Information Technology, TU Chemnitz, Chemnitz, Germany, jingrui.yu@etit.tu-chemnitz.de}
\affil[2]{\protect\raggedright
  Faculty of Electrical Engineering and Information Technology, TU Chemnitz, Chemnitz, Germany, roman.seidel@etit.tu-chemnitz.de}
\affil[3]{\protect\raggedright 
  Faculty of Electrical Engineering and Information Technology, TU Chemnitz, Chemnitz, Germany, g.hirtz@etit.tu-chemnitz.de}
	
\abstract{We propose a one-step person detector for top-view omnidirectional indoor scenes based on convolutional neural networks (CNNs). While state of the art person detectors reach competitive results on perspective images, missing CNN architectures as well as training data that follows the distortion of omnidirectional images makes current approaches not applicable to our data. The method predicts bounding boxes of multiple persons directly in omnidirectional images without perspective transformation, which reduces overhead of pre- and post-processing and enables real-time performance. The basic idea is to utilize transfer learning to fine-tune CNNs trained on perspective images with data augmentation techniques for detection in omnidirectional images. We fine-tune two variants of Single Shot MultiBox detectors (SSDs). The first one uses Mobilenet v1 FPN as feature extractor (moSSD). The second one uses ResNet50 v1 FPN (resSSD). Both models are pre-trained on Microsoft Common Objects in Context (COCO) dataset. We fine-tune both models on PASCAL VOC07 and VOC12 datasets, specifically on class person. Random 90-degree rotation and random vertical flipping are used for data augmentation in addition to the methods proposed by original SSD. We reach an average precision (AP) of 67.3\,\% with moSSD and 74.9\,\% with resSSD on the evaluation dataset. To enhance the fine-tuning process, we add a subset of HDA Person dataset and a subset of PIROPO database and reduce the number of perspective images to PASCAL VOC07. The AP rises to 83.2\,\% for moSSD and 86.3\,\% for resSSD, respectively. The average inference speed is 28\,ms per image for moSSD and 38\,ms per image for resSSD using Nvidia Quadro P6000. Our method is applicable to other CNN-based object detectors and can potentially generalize for detecting other objects in omnidirectional images.}

\keywords{Convolutional Neural Networks (CNNs), Transfer Learning, Omnidirectional Images, Fisheye Camera, Object Detection, Active Assisted Living (AAL)}

\maketitle

\section{Introduction}
\label{sec:intro}

Convolutional neural networks (CNNs) were successfully investigated for several tasks in computer vision in the recent years.
The detection of objects in images belongs to these tasks.
A main requirement for the detection of objects in images for current CNNs are accurate real-world training data.
In this paper we propose a method to detect objects in fish-eye images of indoor scenes using a state-of-the-art object detector.
The object detection in indoor scenes with a limited number of image sensors can be reached with images from omnidirectional cameras.
These cameras are suited for capturing one room with a single sensor due to a field of view of about \ang{180}.
Our goal is to detect objects in indoor scenes in omnidirectional image data with a detector trained on a mixture of perspective and omnidirectional images.
Due to the lack of annotated training data for objects omnidirectional images we choose public available perspective datasets \cite{pascalvoc, coco} and sparsely available omnidirectional images with bounding box ground truth \cite{hda, bomni, piropo}.

The application of our work is the behaviour analysis of elderly with incipient dementia that focuses the object detection on the person class.
To analyse the test persons' behaviour we need the exact position of the person in their own flat.
For this the sensor is mounted on the ceiling of each room to cover the whole living area.

Beside our application, the field of active assisted living, the detection of objects in omnidirectional image data can be used in mobile robots and in the field of autonomous driving.

The remainder of this paper is structured as follows:
Section \ref{sec:related_work} presents previous research activities in object detection in omnidirectional images.
In section \ref{sec:method} we discuss the usage of datasets for training, introduce our own dataset and describe the model architecture and the training itself.
Section \ref{sec:experimentalresults} explains our experiments, the evaluation and performance estimation on state of the art hardware.
Section \ref{sec:conclusion} summarizes the paper's content, concludes our observations and gives ideas for future work.

\section{Related Work}
    \label{sec:related_work}
    \textbf{Person Detection in Omnidirectional Images} A direct approach for detecting objects in omnidirectional images without CNNs was shown in the work of \cite{cinaroglu2014direct}.
    The classical HoG features and training a SVM to detect humans in a transformed INRIA dataset leads to competitive results in recall and precision.
    A novel model named Past-Future Memory Network (PFMN) was proposed by \cite{lee2018memory} on \ang{360} videos.
    One of the main contributions of \cite{lee2018memory} is to learn the correlation between input data from the past and future.
    In contrast to our work, the authors of Spherical CNN \cite{2018spherical} modify the architecture  of ResNet.
    Their goal is to build a collection of spherical layers which are rotation-equivariant and expressive.
    The detection of persons in virtual perspective views \cite{findeisen2013fast} and back transformation to the omnidirectional image source was done by \cite{seidel2018improved}.
    With the goal to find objects in a special area and the application field of data protection, the authors of \cite{callemein2018low} train a state of the art detector on persons and test on equally low resolution images.
    In contrast to this work, we use different person datasets to avoid overfitting in the model to a certain indoor environment.

\section{Method}
\label{sec:method}
  We describe the datasets for training and test and introduce our own dataset in section \ref{sec:dataprep}. Models used for OmniPD and details of the training are discussed in section \ref{sec:training}.

  \subsection{Data Preparation}
  \label{sec:dataprep}
    We use data from multiple public datasets as well as our own dataset to construct the final datasets for training and test. 

    \textbf{PASCAL VOC dataset \cite{pascalvoc}:} images that contain the class person are extracted from the original dataset from 2007 train and test set and 2012 train set.
    A total of 4192 images from 2007 train and test set and 9583 images from 2012 train set are extracted for training our OmniPD. 

    \textbf{HDA Person Dataset \cite{hda}:} only images from \textit{Cam 02} are used for the sake of the omnidirectional viewing angle of this subset. 
    The bounding boxes provided by the authors are slightly misaligned for some of the images. These are manually corrected. 
    Frames without objects are excluded from the dataset. The dataset size is 1388 images.

    \textbf{PIROPO Database \cite{piropo}:} we choose the training sets to enhance our training, which are in detail \textit{training}, \textit{training\_center}, \textit{training\_seat1} and \textit{training\_seat2} in the PIROPO Database. 
    The original datasets are annotated with points marking the positions of persons in the images, which does not suffice for our training. 
    Therefore we manually annotate the images with bounding boxes. 
    The original image sequences are down sampled by a factor of 5 to reduce similarities between the images and to lower the amount of manual labor of annotation. 
    This also reduces the dataset size to 7229 images, which is more comparable to the size of our extracted PASCAL VOC person dataset, eliminating the influence of data imbalance in the training.

    \textbf{Bomni-DB \cite{bomni}:} the database is annotated with bounding boxes only for the moving persons since it is built for the purpose of tracking. 
    We add annotations for the stationary persons in the scenes. 
    We use only images from the top-mounted camera and down sample the sequences by a factor of 10, resulting in a dataset of 1034 images.

\begin{figure}
      \includegraphics[width=0.48\columnwidth]{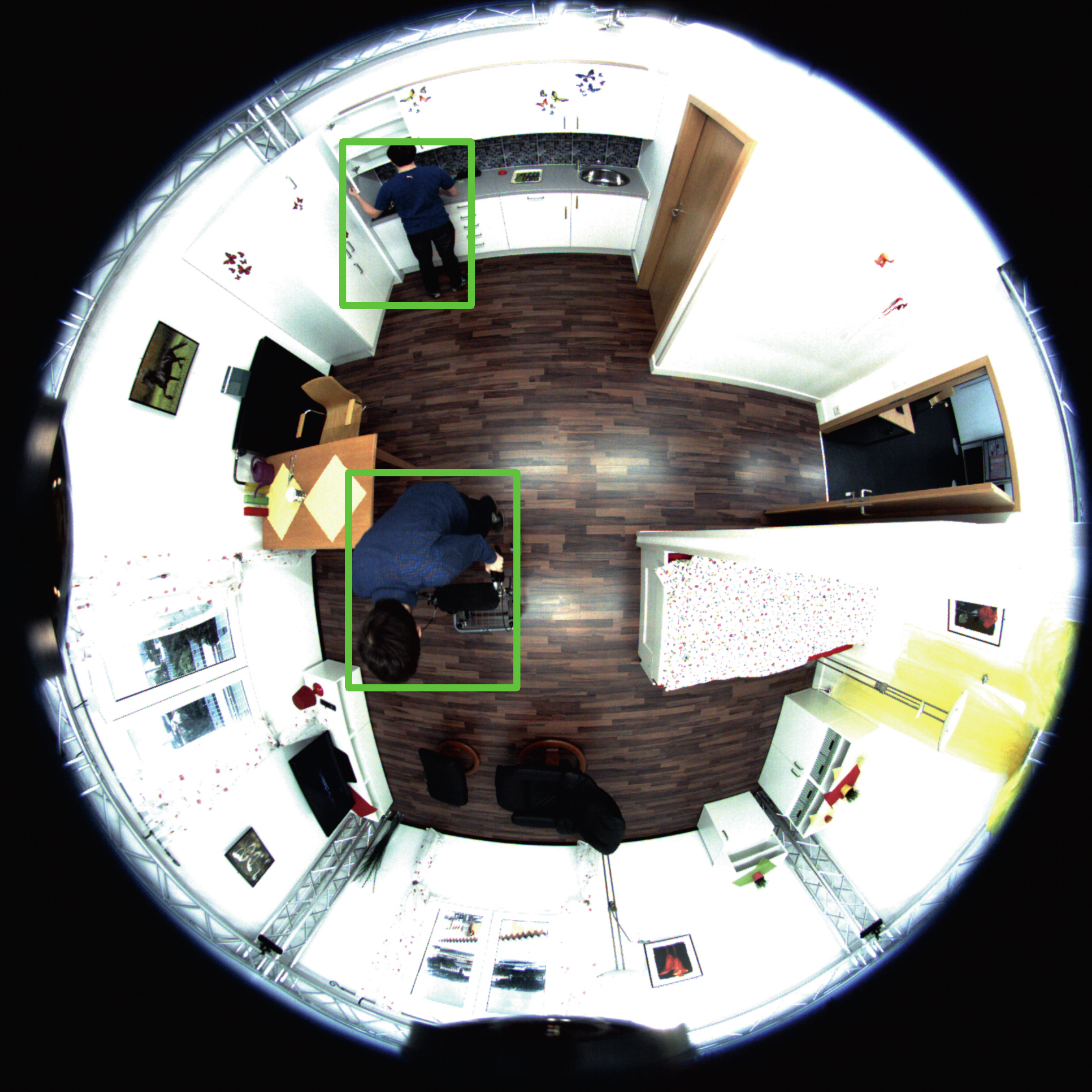}
      \hfill
      \includegraphics[width=0.48\columnwidth]{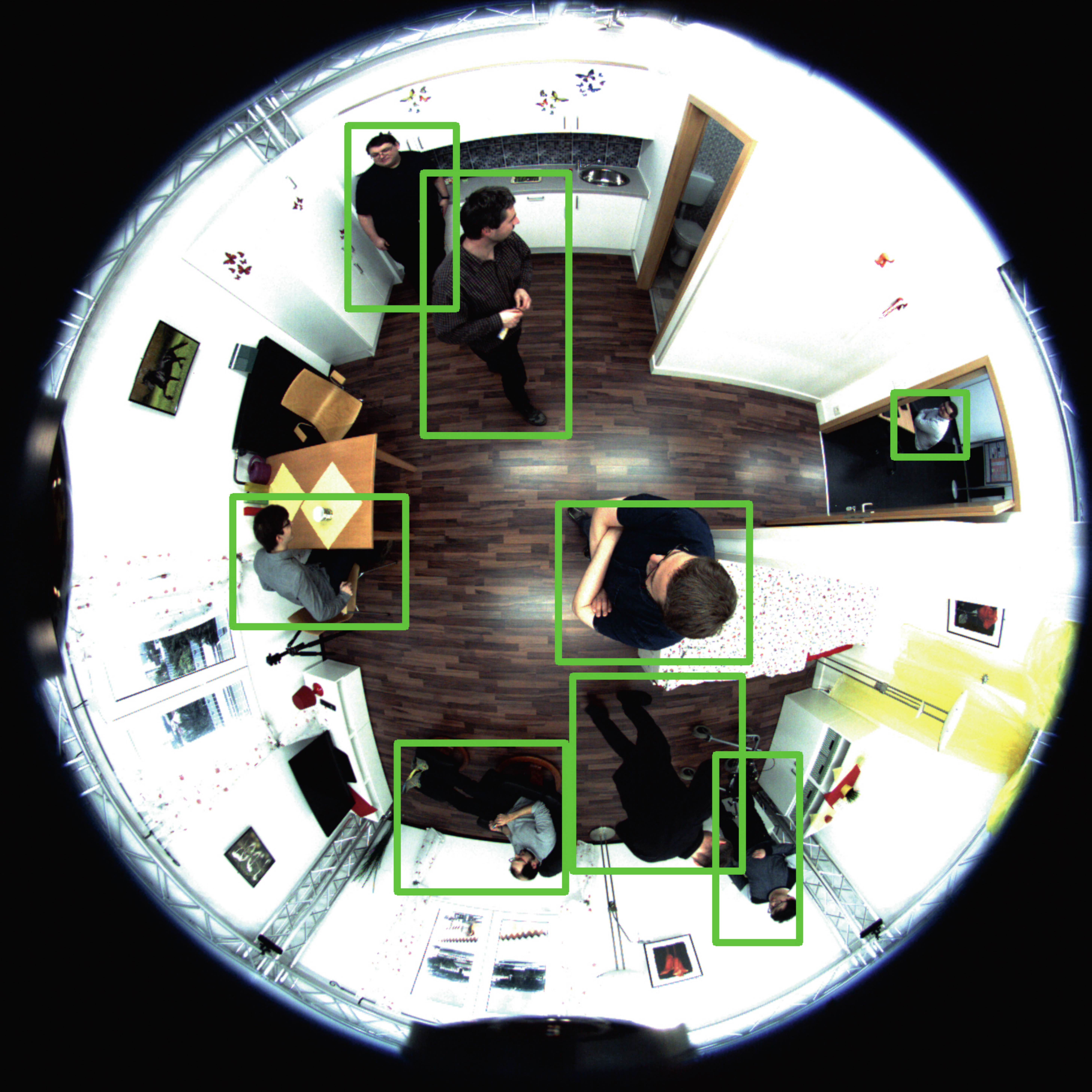}
    	\caption{Sample images with annotations from the DST dataset.}
    	\label{img:dstsample}
    \end{figure}

    \textbf{DST Dataset:} we create our own dataset in an experimental set-up simulating an apartment for elderly with beginning dementia living on his or her own. 
    We record two sequences with 400 and 301 images each. 
    In the first sequence, 3 persons walk around in the apartment, with at most 2 of them in it at the same time. 
    The second sequence features a crowded scene with up to 8 persons in different poses in the apartment at the same time. 
    Bounding boxes are drawn for all persons. Sample images are shown in Figure \ref{img:dstsample}.

    The detailed dataset combinations for training and test are listed in Table \ref{tab:datasets}. VOC07 stands for the combined PASCAL VOC 2007 train and test set. 

    \begin{table}[h]
      \caption{Datasets for training and test for class person.}
      \begin{tabular}{lcccc}
        & train\_voc & train\_hp & train\_hpv07 & test\_db \\ \midrule
      VOC07 & \checkmark & & \checkmark & \\
      VOC12 train & \checkmark & & & \\
      HDA Cam 02 & & \checkmark & \checkmark & \\
      PIROPO train & & \checkmark & \checkmark & \\
      Bomni-DB & & & & \checkmark \\
      DST (ours) & & & & \checkmark \\ \midrule
      Dataset Size & 13775 & 8567 & 12759 & 1735 \\
      \end{tabular}
      \label{tab:datasets}
    \end{table}

  \subsection{Model Selection and Training}
  \label{sec:training}
    Two variants of Single Shot MultiBox Detector (SSD) \cite{ssd} are chosen from Tensorflow detection model zoo \citep{tf-od, tfzoo} based on their inference speed and accuracy, which is represented with their mean average precisions (mAP) on COCO 2014 minival dataset \cite{coco}. 
    SSD with Mobilenet V1 FPN \citep{mobilenet,fpn} as feature extractor (moSSD) is lightweight and fast, which achieves the mAP of 32\,\% at 56\,ms per image on Nvidia Titan X GPU. 
    SSD with ResNet50 V1 FPN \cite{resnet} (resSSD) achieves better mAP of 35\,\% at 76\,ms per image \cite{tfzoo}. 

    Both SSD models are fine-tuned from the weights trained on COCO dataset on 3. Jul. 2018 provided in Tensorflow detection model zoo. 
    We use pre-defined training protocols from Tensorflow Object Detection API \cite{tf-od} for the respective models. 
    Aside from random horizontal flipping, random colour distortion, random RGB to grey conversion and random image cropping, random vertical flipping and random 90-degree rotation are added to data augmentation methods. 
    Both models are fine-tuned on all three training datasets with all layers open for update. 
    We then lock layers of the feature extractor (FE) for both models and train them on \textit{train\_hpv07} for comparison. 
    Finally we fine-tune all layers of R-FCN with ResNet101 trained on COCO dataset on 1. Jan. 2018 provided in Tensorflow detection model zoo to verify that our method is applicable to other network structures than SSD. 
    We modify learning rate and number of epochs for training as provided in Table \ref{tab:training}. 
    A cosine learning rate decay curve \cite{loshchilov2016sgdr} is applied to all trainings.

    \begin{table}[h]
      \caption{Learning rate and epochs of training.}
      \begin{tabular}{lcc}
        Model + Training Dataset & Learning Rate & Epochs \\ \midrule
        moSSD + train\_voc & 0.002 & 40 \\
        moSSD + train\_hp & 0.002 & 40 \\
        moSSD + train\_hpv07 & 0.002 & 25 \\
        moSSD (FE locked) + train\_hpv07 & 0.002 & 35 \\
        resSSD + train\_voc & 0.0005 & 45 \\
        resSSD + train\_hp & 0.0005 & 45 \\
        resSSD + train\_hpv07 & 0.0005 & 30 \\
        resSSD (FE locked) + train\_hpv07 & 0.0005 & 30 \\
        R-FCN + train\_hpv07 & 0.00008 & 40 \\ 
      \end{tabular}
      \label{tab:training}      
    \end{table}

\section{Experimental Results}
\label{sec:experimentalresults}  
  We present the evaluation results of the trained models on the test dataset and analyse them in section \ref{sec:evaluationresults}. 
  We also test the inference speed of the models and present the results in section \ref{sec:temporalperformance}.

  \subsection{Evaluation Results}
  \label{sec:evaluationresults}
    We use the mean average precision (mAP) at the intersection over union (IoU) value of 0.5 defined in the PASCAL VOC \cite{pascalvoc}. 
    Since we only have one class person, the mAP@0.50\,IoU is the same as AP\,(person)@0.50\,IoU. 
    The evaluation results are shown in Table \ref{tab:results}.

    \begin{table}[hbt]
      \caption{Evaluation results on \textit{test\_db} given in mAP@0.50\,IoU (\%) / Epoch with best results.}
      \begin{tabular}{llll}
        Training Dataset & moSSD & resSSD & R-FCN \\ \midrule
        train\_voc & 67.3 / 26 & 74.9 / 32 & \\
        train\_hp & 69.2 / 19 & 71.6 / 30 & \\
        train\_hpv07 & 83.2 / 24 & 86.3 / 21 & 84.4 / 22 \\
        train\_hpv07, FE locked & 69.3 / 31 & 74.0 / 26 & \\
      \end{tabular}
      \label{tab:results}
    \end{table}

    As presented in Table \ref{tab:results}, the trained models reach their highest mAP already before the training ends. The reason is overfitting. 
    Since we have only one class with around 10,000 images in each training dataset, while the models are designed for larger tasks like COCO or PASCAL VOC, their capacities are much larger than what our data can provide. 
    Towards the end of the training, the model severely overfits to the training dataset, resulting in a lower generalization ability. 
    This leads to a decay of detection mAP, as our test dataset is completely decoupled from the training datasets.

    The results are similar for both SSD models, with resSSD performing slightly better than moSSD, which is in consistency with the reference mAP given in Tensorflow detection model zoo. 
    We focus on the differences between training datasets. 

    As expected, training solely on perspective images from PASCAL VOC dataset delivered moderate detection mAP at 67.3\,\% and 74.9\,\%. 
    The training converged faster with similar mAPs for both models after switching to \textit{train\_hp} due to smaller variances in the dataset. 
    Combining perspective images and omnidirectional images for training improve detection performance by a large margin. 
    The absolute average improvement over the four previous experiments is 14\,\%. moSSD and resSSD reach the mAP of 83.2\,\% and 86.3\,\%, respectively.
    Locking the layers in the feature extractor lowered the mAP back to the previous level.


    The R-FCN model trained on \textit{train\_hpv07} achieved similar performance as the SSDs, which proves that our method is applicable to other CNN-based object detectors.

  \subsection{Temporal Performance}
  \label{sec:temporalperformance}
    We measure the speed of the trained models on a workstation with Intel Core i7-6900K CPU, 32GB RAM and Nvidia Quadro P6000 GPU. 
    We restrict the output image resolution of our camera to 640$\times$640 pixels to match the input size of the SSDs. 
    Using the frozen inference graph for detection, moSSD takes 28\,ms per image and resSSD takes 38\,ms per image. 
    Both models achieve stable real-time detection speed of over 20 frames per second with the overhead of pre- and post-processing taken into account, which is about 5.3\,ms. 
    The inference time for R-FCN is 46\,ms per image.

\section{Conclusion}
  \label{sec:conclusion}
  In this work we successfully train CNN-based object detectors for person detection in omnidirectional images of indoor scenes. 
  The trained models moSSD, resSSD and R-FCN achieve detection APs for the class person of 83.2\,\%, 86.3\,\% and 84.4\,\% on our test dataset without pre- and post-processing specific to omnidirectional images. 
  This performance is on par with the detection AP for class person of leading edge CNN detectors such as DSSD at 86.4\,\% \cite{dssd}, which is tested on PASCAL VOC 2012 test set. 
  We manage to maintain real-time performance for all three models at the same time. 
  The main contributions of our work are three-fold:
  \begin{enumerate}
    \item we enhance existing datasets of omnidirectional images from top-view cameras and create an independent test dataset for person detection that share no common sources with the training dataset. 

    \item we combine images of normal human perspective and images directly from top-mounted fisheye cameras for training to solve the two common problems of using CNNs for object detections in omnidirectional images: \begin{itemize}
      \item lack of training data in the application domain,
      \item feature mismatch due to perspective differences and camera distortion.
    \end{itemize}
    The feature mismatch is proven by the evaluation results of models with locked feature extractor.
    
    \item We eliminate overhead from perspective transformation by directly applying CNN-based detectors on omnidirectional images.
  \end{enumerate}

  To our knowledge this is the first time CNN-based detectors are successfully applied to omnidirectional images without adaptations to accommodate the distortion and the special view angle of top-mounted fisheye cameras. 
  Our method can still be further enhanced by using more advanced data augmentation methods, such as four-point transformation on normal perspective images to mimic the distortion of omnidirectional images.
  Synthetic data can be generated with virtual fisheye cameras for the training. 
  It is also feasible to use our method to train detectors for other objects besides person.

\begin{acknowledgement}
  This project is funded by the European Regional Development Fund (ERDF) and the Free State of Saxony under the grant number 100-241-945. We also thank NVIDIA for providing the GPU used for training the networks.
\end{acknowledgement}

\bibliographystyle{unsrt}
{\small
\bibliography{lit}}

\end{document}